\documentclass{ecai} 

\usepackage{color}
\usepackage{dsfont}
\usepackage{multirow}
\usepackage{xspace}
\usepackage{booktabs}
\usepackage{hyperref}
\usepackage{mdframed}
\usepackage{siunitx}
\usepackage[abbreviations]{foreign}
\usepackage{enumitem}
\usepackage{breakurl}

\usepackage{amsmath,amssymb,amsfonts, amsthm, mathtools, stmaryrd}
\usepackage{algorithmic}

\usepackage[bottom]{footmisc}

\usepackage{scalerel} 

\usepackage{graphicx}
\usepackage{subcaption}
\usepackage{textcomp}
\usepackage{xcolor}
\def\BibTeX{{\rm B\kern-.05em{\sc i\kern-.025em b}\kern-.08em
    T\kern-.1667em\lower.7ex\hbox{E}\kern-.125emX}}

\usepackage{pifont}

\usepackage{cleveref}

\theoremstyle{plain}
\newtheorem{theorem}{Theorem}[section]

\theoremstyle{definition}
\newtheorem{definition}[theorem]{Definition}

\theoremstyle{remark}

\usepackage{thmtools}
\usepackage{thm-restate}

\newcommand{\restatableeq}[3]{\label{#3}#2\gdef#1{#2\tag{\ref{#3}}}}

\newcommand{\apriori}{a-priori\xspace}
\newcommand{\aposteriori}{a-posteriori\xspace}
\newcommand{\selfish}{no collab.\xspace}

\newcommand{\unif}{uniform\xspace}
\newcommand{\stratified}{stratified\xspace}
\newcommand{\neyman}{Neyman\xspace}

\newcommand{\task}{ACSPublicCoverage\xspace}
\newcommand{\gender}{\texttt{SEX}\xspace}

\newcommand{\nativity}{\texttt{NATIVITY}\xspace}

\newcommand{\mig}{\texttt{MIG}\xspace}

\newcommand{\agep}{\texttt{AGEP}\xspace}

\newcommand{\mar}{\texttt{MAR}\xspace}
\newcommand{\gc}{German Credit\xspace}
\newcommand{\pp}{Propublica\xspace}
\newcommand{\ft}{Folktables\xspace}
\newcommand{\var}[1]{Var#1}

\usepackage{acronym}
\acrodef{DP}{demographic parity}
\acrodef{ML}{machine learning}
\acrodef{FL}{Federated Learning}

\begin{document}

\begin{frontmatter}

\paperid{2287}

\title{Fairness Auditing with Multi-Agent Collaboration}

\author[A, *]{\fnms{Martijn}~\snm{de Vos}
}
\author[A, *]{\fnms{Akash}~\snm{Dhasade}
}
\author[B, *]{\fnms{Jade}~\snm{Garcia Bourrée}
}
\author[A]{\fnms{Anne-Marie}~\snm{Kermarrec}
}
\author[B]{\fnms{Erwan}~\snm{Le Merrer}
}
\author[C]{\fnms{Benoît}~\snm{Rottembourg}
}
\author[D]{\fnms{Gilles}~\snm{Tredan}
}

\address[A]{EPFL, Lausanne, Switzerland}
\address[B]{Inria, University of Rennes, Rennes, France}
\address[C]{Inria, Paris, France}
\address[D]{LAAS, CNRS, Toulouse, France}
\address[*]{Corresponding Authors. Email: martijn.devos@epfl.ch, akash.dhasade@epfl.ch, jade.garcia-bourree@inria.fr}

\begin{abstract}
    Existing work in fairness auditing assumes that each audit is performed independently.
    In this paper, we consider multiple agents working together, each auditing the same platform for different tasks.
    Agents have two levers: their collaboration strategy, with or without coordination beforehand, and their strategy for sampling appropriate data points.
    We theoretically compare the interplay of these levers.
    Our main findings are that \emph{(i)} collaboration is generally beneficial for accurate audits, \emph{(ii)} basic sampling methods often prove to be effective, and \emph{(iii)} counter-intuitively, extensive coordination on queries often deteriorates audits accuracy as the number of agents increases.
    Experiments on three large datasets confirm our theoretical results.
    Our findings motivate collaboration during fairness audits of platforms that use ML models for decision-making.
    
\end{abstract}

\end{frontmatter}


\section{Introduction}
\Ac{ML} models are becoming an integral part of many business and industrial processes, increasingly impacting various facets of our lives~\citep{holstein2019improving}.
Such models are increasingly employed to drive decisions in high-stakes domains~\cite{buyl2024inherent}.
For example, many financial institutions use AI-driven models in which several attributes such as income, credit score, and employment history influence the decision to issue a particular loan~\cite {mhlanga2021financial}.
These models are also used to automate the hiring process of certain companies, which would otherwise be labor-intensive~\cite{giang2018potential,hunkenschroer2022ethics}.
Because these models may significantly impact people's lives, their fairness and regulatory compliance have become increasingly important~\cite{pessach2022review, le2023algorithmic}.

Estimating the fairness of \ac{ML} models is commonly done through algorithmic audits by regulators~\cite{ng2021auditing}.
However, auditors are not granted unrestricted access to a \ac{ML} model to protect trade secrets but instead send queries to the model and use the query responses (\textit{i.e.}, a \textit{black-box} interaction) to detect fairness violations.
It is common to impose a cap on the queries that are sent to the black-box in order not to overload or interfere with the model being audited~\citep{neiswanger2021bayesian, yan2022active, rastegarpanah2021auditing}.

As of today, an auditor performs her audit tasks on each attribute of interest sequentially, one at a time, and independently of other auditors. For example, if she wants to audit a bank's \ac{ML} model that predicts whether it is safe to issue a loan~\cite{feldman2015certifying}, she begins by auditing the fairness property of the \emph{gender} attribute in a first step. In a subsequent step, she independently audits the fairness property on the \emph{race} attribute. This procedure could result in a sub-efficient auditing scheme in terms of the amount of queries sent to the model. Instead, a coordinated --or \textit{collaborative}-- auditing scheme, in which information is shared between distinct audits, might have been more effective. In other words, there is an opportunity to mutualize queries, \ie through collaboration between the different agents of an auditor. While collaboration in a \textit{single} audit task has recently been introduced in the community~\cite{AIStatsFAIR}, to the best of our knowledge, collaborative auditing for multiple tasks has not yet been studied. Therefore, we pose the question: \emph{can an auditor benefit from collaboration among individual audit tasks, e.g., by strategically constructing and sharing queries and responses}?

We answer this question by studying collaboration strategies for independent auditing agents. Specifically, their common goal is to enhance the efficiency and accuracy of auditing a target model for one of the most studied fairness estimation tasks: \emph{\ac{DP}}~\cite {yan2022active,singh2023measures,pessach2022review}. To this end, we introduce and analyze two realistic forms of multi-agent collaboration. In the \emph{\aposteriori} collaboration, agents share their queries and the responses they receive. In the \emph{\apriori} collaboration, in addition, agents coordinate beforehand on their queries to maximize the information that can be gathered. Besides the types of collaboration, auditors need to wisely choose the strategy for querying data points for estimating \ac{DP}. These strategies are the sampling methods that address the model's input space that are suitable to the audit black-box setup. Thus, the scientific challenge of this paper is to analyze the relevant combinations of collaboration strategies and sampling methods.

\textbf{Contributions.} This work makes the following contributions:
\begin{enumerate}[noitemsep,topsep=0pt]
    \item We propose a multi-agent setup in which agents collaboratively perform fairness audits of \ac{ML} models (\Cref{s:se-G}).
The collaborations are driven using coordinated sampling methods on sensitive attributes under audit and by sharing query responses.
    \item We provide a theoretical analysis of the effectiveness of the \apriori and \aposteriori collaboration strategies and their interplay with different sampling methods (\Cref{s:th-B}).
    First, we show that collaboration is generally beneficial for audit accuracy compared to conducting independent audits.
    Second, we derive that the advantages of sampling strategies vanish when the number of auditors increases for \aposteriori collaboration.
    Third, we show that, surprisingly, performing extensive coordination on queries when the number of agents increases sometimes \emph{hurts} audit accuracy.
    \item Using three real-world large datasets (\ft, \gc, and \pp), commonly used in fairness studies, we empirically confirm our main theoretical findings (\Cref{s:exp}).
\end{enumerate}

This work is the first to explore the nuances and effectiveness of collaboration between different fairness audit tasks of black-box \ac{ML} models.
In summary, we find that collaboration among agents is a successful setup for increasing query efficiency and detecting biases.

\textbf{Related Work.}
We refer the reader to surveys such as~\cite{pessach2022review, caton2020fairness, mehrabi2021survey} for a general introduction to fairness.
While the predominant focus within this domain lies on the fair learning, existing work extends to various subjects,
including auditing black-box settings~\cite{agarwal2018reductions, maneriker2023online}.

The exploration of fairness in multi-agent systems using game-theoretic frameworks has received relatively little attention~\cite{de2008fairness}. Previous work on fairness in collaborative frameworks, like FAIR~\cite{AIStatsFAIR}, emphasizes data sharing and fairness between agents, but it does not align with the objective of estimating fairness in specific black-box models.
FAIR focuses on fair collaboration between agents with different devices for scientific discovery, while we focuse on agents of equal importance studying the fairness of a common algorithm.

Beyond technical considerations, legal dimensions also influence the fairness landscape.
Traditional legal protocols often constrain inter-agency collaborations for assessing legal compliance, with exceptions such as the recent precedent set by~\cite{eu2522023}.
\cite{cremer2023enforcing} examines the European Commission's implementation of the Digital Markets Act~\cite{DMA} in March 2024 and offers valuable insights and recommendations for optimizing compliance mechanisms and resolving Big Tech platform investigations effectively, along with recommendations for collaborative processes.
 
In conclusion, we expect our work to facilitate effective collaboration between different agencies, streamlining regulatory efforts and enhancing law enforcement.

\section{Background: Sampling}
\label{ss:sampling}

Let us consider an auditor that wants to know the average value $\mu$ of a particular numerical $\{x_i\}_{1\leq i\leq N}$ value in a population of size $N$, \ie, $\mu = \frac{1}{N}\sum_{i=1}^Nx_i$. In situations where this auditor can only afford consulting a fraction $R<N$  of such values, she has to resort to an \emph{estimator} $\hat{\mu}$.
The Bienaymé-Chebyshev inequality states that $\hat{\mu}$, a random variable, has finite variance $Var$, the probability that $\hat{\mu}$ deviates from its mean by more than a standard deviation is bounded by $Var$. We hence use variances to characterize empirical errors.

Besides analyzing $\hat{\mu}$ for different sampling methods, this section also highlights the sampling variance of $\hat{\mu}$, which is important as a large sampling variance can lead to inaccurate estimates.

This work explores three sampling methods that auditors use to construct queries: \emph{(i)} uniform sampling, \emph{(ii)} stratified sampling, and \emph{(iii)} Neyman sampling~\cite{arieska2018margin, mathew2013efficiency}.
While uniform sampling is a natural, well-studied, and easy-to-implement choice, stratified sampling offers a fairer method that an auditor can use in a black-box setting.
Neyman sampling will serve as an upper bound, as it is the most accurate for an "omniscient" auditor (that would have white-box-like information on the problem).

\subsection{Uniform Sampling}
The most straightforward approach to estimate $\mu$ is to select $R$ members of the population randomly.
We refer to the sampled members as $x_1, \dots x_R$.
With uniform sampling, the allocation of queries mirrors the real-world distribution of the population.
The estimate of $\mu$ with uniform sampling, referred to as $ \hat{\mu}_{\unif} $, is the mean of $x_i$ over the sampled set: $\hat{\mu}_{\unif} = \frac{1}{R}\sum_{i=1}^R x_i.$

\textbf{Sampling Variance.} The sample variance associated with uniform sampling is: $\var(\hat{\mu})_{\unif} = \frac{1}{R}\sum_{i=1}^R (x_i - \hat{\mu})^2.$

\subsection{Stratified Sampling}
Uniform sampling often falls short of providing the most accurate estimator, especially in heterogeneous populations. 
Stratified sampling involves dividing the heterogeneous population into $n$ subgroups, called \emph{strata}, such that subgroups are non-overlapping and homogeneous with respect to $\mu$.
Once the population is segmented into strata, one selects $R_j$ samples from each subgroup $ j $. All the samples drawn from each stratum constitute a stratified sample of total size $R = \sum_{j=1}^n R_j$.
Stratified sampling can enhance precision by ensuring that all subgroups are adequately represented, making it a more effective method when dealing with diverse groups~\cite{arieska2018margin, keramat1998study}.

We consider in this work \emph{disproportionate stratified sampling} which is a particular type of stratified sampling and in which an equal amount of the query budget is allocated to each stratum.
Specifically, with $ n $ strata, $R_j = R/n$ of the budget is spent on each stratum $ j $.
This sampling strategy ensures that each stratum of a given attribute is sampled using the same number of queries.

\paragraph{Sampling Variance.} $ \hat{\mu}_{\stratified} $ is given by a weighted average of the estimators by stratum sizes: $\hat{\mu}_{\stratified} = \sum_{j=1}^n p_j \hat{\mu}_j,$ with $p_j$ being the probability to be in the stratum $j$ and $\hat{\mu}_j$ being the estimator of $\mu$ in this stratum with $R_j$ samples.

It is possible to show that the variance under stratified sampling is~\cite{mathew2013efficiency}:
$\var(\hat{\mu})_{\stratified} = \sum_{j=1}^n p_j^2 \left( \frac{1}{R_j} - \frac{1}{p_jN} \right) \var(\hat{\mu}_j)^2$,
with $\var(\hat{\mu}_j) = \frac{1}{R_j}\sum_{i=1}^{R_j} (x_{ji} - \hat{\mu}_{j})^2$ being the sample standard deviation of stratum $j$.

While disproportionate stratified sampling offers advantages in terms of representation, it might not be the optimal strategy.
In particular, in situations where strata are heavily unbalanced w.r.t. their sizes, the resulting strata may be of uneven interest to an auditor, and a more nuanced sampling strategy might have been preferable.

\subsection{Neyman Sampling}
\label{subsec:neyman_sampling}

Neyman sampling is the optimal sampling strategy defined as the stratified sampling strategy in which the allocation of queries among strata yields the most precise estimate of $\mu$~\cite{mathew2013efficiency}.
It minimizes the variance in the estimation process:
$\left(R_1^*, \dots, R_n^*\right) = argmin~\left(\var(\hat{\mu})_{\stratified}\right)$.
Since Neyman sampling is a specific instance of stratified sampling, its mean and variance estimates are identically constructed:
$\hat{\mu}_{\neyman} =  \sum_{j=1}^n p_j \hat{\mu}_j$ and $\var(\mu)_{\neyman} = \sum_{j=1}^n p_j^2 \left( \frac{1}{R_j} - \frac{1}{p_jN} \right) \var(\mu_j)^2$.

However, to compute the Neyman sampling allocation, one has to know the standard deviation values of each stratum beforehand. This assumption is unlikely to be met in practice, as an auditor knowing those values could directly derive $\mu$ values.
Nonetheless, despite its lack of realism, Neyman sampling serves as an optimal baseline to compare practical approaches against it and reveal the impact of auditor's missing knowledge on the precision. 

\textbf{Discussion.} The exploration of sampling methodologies reveals a spectrum of approaches, each with strengths and considerations. Uniform sampling is simple to implement
but may not accurately represent unbalanced populations. Disproportionate stratified sampling ensures fair representation but can also be limited in unbalanced populations. Neyman sampling, although impractical to implement, is considered optimal as it balances stratum size and intra-stratum variance to minimize estimation variance and maximize precision. 
In this paper, we emphasize the importance of choosing a sampling method tailored to population characteristics for reliable results and explore the interaction between these methods and collaborative strategies.

\section{Problem Statement: Collaborative Auditing}
\label{s:background}

To provide a structured framework for our investigation, we formalize the audit process and delineate its key components. 

\textbf{Auditing Fairness.}
We investigate a \textit{black-box} algorithm $\mathcal{A}~:~\mathcal{X}~\mapsto~\{0,1\}$ (\eg, a \ac{ML} model) deployed within a platform setting. The input space $\mathcal{X}$ encompasses a set of $ m $ protected, binary attributes denoted as $X_1, X_2, \dots, X_{m}$.
Most of the work on fairness in ML deals with binary classification problems, and we also adopt this scenario for the sake of consistency~\cite{castelnovo2022clarification, agarwal2018reductions, singh2023measures}.
While, most fairness metrics for binary targets can be generalized to support multi-class classification~\cite{denis2021fairness}, this is beyond the scope of this paper.

In our context, $ m $ distinct agents $A_1, \dots, A_m $ interact with the same black-box algorithm $\mathcal{A}$ to scrutinize fairness attributes associated with specific attributes. Each agent $A_i$ concentrates on assessing a distinct protected attribute $X_i$.
For example, in a loan attribution application~\cite{feldman2015certifying}, some agent $A_1$ could audit a fairness property of the \emph{gender} attribute ($X_1$) and $A_2$ could audit a fairness property on the \emph{race} attribute ($X_2$). 
Each attribute $X_i$ induces two groups in $X$: the favored and unfavored groups. We call a \emph{stratum} each intersection of these $2m$ groups, \ie, with $ m $ agents, there are $2^m$ strata.

In line with related work, we assume that an auditor can issue a fixed total number of $B$ queries, a common preamble in auditing \citep{yan2022active}. 
$ B $ is called the \emph{query budget}.
We assume that each agent can send $ R $ queries to $ \mathcal{A} $ where $ R = B/ m $ is the per-agent query budget. 

\textbf{Independence Assumption.}
The attributes $X_1, X_2, \dots, X_{m}$ are considered \textit{protected} due to their sensitive nature and potential implications for fairness.
In line with other works~\cite{lundberg2017unified, vstrumbelj2014explaining, mougan2023demographic}, we assume that these attributes are \textit{independent} of one another, \ie, the value of one attribute does not influence or depend on the values of other ones.
In theoretical analysis, the assumption of independence among protected attributes is crucial for clarity and insights. However, real-world scenarios may involve interdependencies among these attributes, which is a topic for future investigation.

\begin{figure*}[ht!]
    \centering
    \includegraphics[width=.9\textwidth]{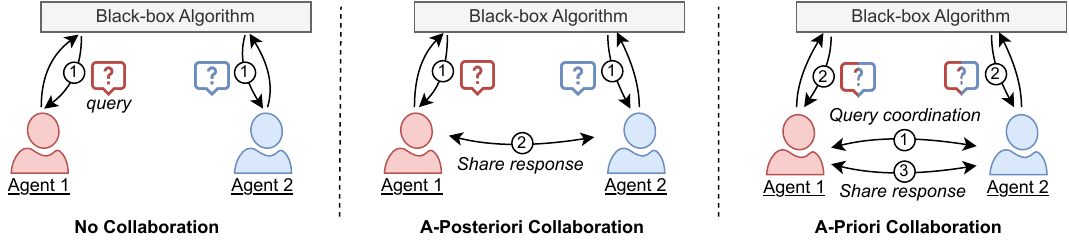}
    \caption{\centering Possible collaboration strategies of an auditor with her two agents: no collaboration (left, baseline), \aposteriori collaboration where agents share queries and responses (middle), and \apriori collaboration where agents also coordinate on queries to be sent (right).}
    \label{fig:framework}
    \vspace{0.3cm}
\end{figure*}

\textbf{Fairness Metric.}
The notion of fairness has been operationalized in several different metrics such as disparate impact, demographic parity, or equalized odds. We refer the reader to the recent survey of Pessach and Shmueli~\cite{pessach2022review} for a comprehensive overview of the topic. 

\Acf{DP}, mathematically denoted as $D$, has emerged as a critical fairness metric for modeling social equality~\citep{buyl2024inherent, mouzannar2019fair, yan2022active}.
An algorithm satisfies \ac{DP} if, on average, it produces the same predictions across different protected groups.
Despite its simplicity, \ac{DP} is a key measure of fairness, especially in high-stakes areas such as finance and recruitment, where decisions can significantly impact individuals' lives.
EU regulations also recognize \ac{DP} as a metric for detecting bias in algorithmic decision-making~\cite{europarl2022}.

\begin{definition} The demographic parity $D_i$ with respect to a binary protected attribute $X_i$ captures the impact of $X_i$ on a prediction $ Y $:\\
$D_i = \mathds{P}(Y = 1 | X_i = 1) - \mathds{P}(Y = 1 | X_i = 0).$
\end{definition}
Agent $A_i$ can use her $R$ queries to estimate \ac{DP} on protected attribute $X_i$ by querying any $x\in \mathcal{X}$.
Let $\hat{D}_i$ be a \ac{DP} estimator:

\begin{align}
\restatableeq{\dpestimation}{\hat{D}_i = \mathds{\hat{P}}(Y = 1 | X_i = 1) - \mathds{\hat{P}}(Y = 1 | X_i = 0)}{dpestimation}.
\end{align}

For brevity, we will write
$\hat{Y}_i = \mathds{\hat{P}}(Y = 1 | X_i = 1)$ and\\ $\hat{Y}_{\bar{i}} = \mathds{\hat{P}}(Y = 1 | X_i = 0)$.

According to the Digital Services Act~\cite{DSA}, if $D_{i} = 0 \pm 0.2$, then $\mathcal{A}$ respects demographic parity on protected attribute $X_i$. While demographic parity could, in addition, be estimated relatively to each stratum (as in intersectional fairness~\cite{10.24963/ijcai.2023/742}), we focus in this paper on attribute-level \ac{DP}.
\ac{DP} is a group level fairness metric according to the classification proposed by Pessach and Shmueli  ~\cite{pessach2022review}, which is defined in opposition to individual-level fairness metrics. 

While our work focuses on~\ac{DP} for concreteness, we argue that any group-level metric requires an auditor to sample members of different target groups to estimate the group property of interest and has to deal with the convergence of empirical estimators. Our work can hence be easily transposed to other group metrics. For instance, disparate impact uses the ratio of group-level estimates rather than their difference for \ac{DP} (see eq.~\ref{dpestimation}).
Hence, while those two estimators will have different variances, the strategy of minimizing the variance of each subgroup is the same.

\textbf{Objective of an Auditor.} The goal of an auditor is to audit \ac{DP} as accurately as possible, \ie, by obtaining a \ac{DP} estimate that is close to the ground truth \ac{DP}, which is the value that can be computed provided one has access to the whole dataset.
An auditor is typically interested estimating the \ac{DP} for different attributes; in this paper, each such attribute estimation is embodied and performed by an \textit{agent}.
We consider the estimations of all attributes as equally valuable to the auditor: the auditor's goal is to have \textit{the average of \ac{DP} variances, i.e., the difference between her estimated \ac{DP}s and the actual \ac{DP}s, as small as possible}.
Note that the choice of a regular average as the metric is arbitrary, acknowledging that alternative metrics exist; for instance, \citet{AIStatsFAIR} proposes a metric ensuring fairness between agents.
However, in our context, where all agents share the common goal of auditing and are controlled by a single auditor, this paper strives for the lowest averaged \ac{DP} variance.

Finally, we assume that agents are homogeneous: they all use the same sampling and collaboration strategy.
We leave the analysis of collaboration among heterogeneous agents to future work.

\section{Multi-Agent Collaboration}
\label{s:se-G}
To estimate \ac{DP}, an agent relies on \emph{(i)} a sampling method that influences the way she constructs her queries, and \emph{(ii)} a collaboration strategy with other agents.
We first introduce two collaboration strategies and then derive the sampling variances of these strategies.

\subsection{Collaboration Strategies}
We introduce two natural collaboration strategies in a black-box audit context and a baseline approach that does not involve collaboration.

\textbf{No Collaboration (baseline).}
To quantify the effectiveness of our collaboration strategies, we consider a non-collaborative setting where each agent queries the black-box algorithm by itself and independently of the others. This baseline is also used in \cite{AIStatsFAIR} and represented in the left part of \Cref{fig:framework}.

\textbf{\textit{\aposteriori} Collaboration.}
The most natural collaboration scheme involves sharing queries and their responses among agents.
In this approach, each agent independently queries the black-box algorithm and then shares the queries and resulting responses with the other agents so that all agents can access a pooled set of queries.
This is also visualized in the middle part of~\Cref{fig:framework}.
Even though a query-response pair obtained by agent $ i $ might not have been optimized to satisfy the interest of agent $ j $, $ j $ might still be able to leverage the information to obtain a more precise \ac{DP} estimation on attribute $ X_j $.

\textbf{\textit{\apriori} Collaboration.}
We introduce a second collaborative strategy, aiming at a preliminary coordination of agents.
With \apriori collaboration, agents collectively implement a sampling strategy, accounting for the estimation tasks of the other participating agents.
This is also visualized in the right part of~\Cref{fig:framework}.
More specifically, all agents divide the input space into $2^m$ strata and agree on the sampling strategy of each stratum.
This coordination allows for a more integrated and strategic approach to querying, intending to enhance the overall effectiveness of the audit.

Note that a-priori and a-posteriori collaboration strategies cover the possible options in our black-box audit setting.
Specifically, there are only two ways to collaborate in a bulk query-response-based approach: before or after sending the queries.
\apriori collaboration has coordination before sending the queries, and \aposteriori shares queries and their responses after they have been sent to the platform.

\subsection{From Collaborations to Estimation Variances}\label{ss:sampling-variance}
The objective of an auditor is to use a budget of $R$ per-agent queries to derive an accurate estimator $\hat{D}_i$ of the demographic parity $D_i$ of their respective protected attribute $X_i$.
Note that $\hat{D}_i$ is a random variable.
Analyzing its distribution is key to understanding the characteristics of the different estimators that result from the interplay of sampling and collaborations employed.
This section thus analyzes the distribution of $\hat{D}_i$ for our collaboration strategies.

As seen in Eq.~\eqref{dpestimation}, the demographic parity, $\hat{D}_i$, is determined by comparing the two empirical probabilities $\hat{Y}_i$ and $\hat{Y}_{\bar{i}}$. Each empirical probability is calculated as the proportion of positive or negative responses collected from queries within the group of interest, \eg,
$\hat{Y}_i = \frac{|X_i = 1, Y = 1|}{|X_i = 1|}.$
The empirical probability $\hat{Y}_{\bar{i}}$ is defined by replacing $X_i = 1$ by $X_i = 0$ in the previous equation.

As the probabilities $Y_i$ and $Y_{\bar{i}}$ can be calculated as the average number of positive answers on the stratum verifying $X_i = 1$, respectively $X_i = 0$, our study inherently accommodates the sampling methodologies presented in~\Cref{ss:sampling}.

By linearity of the variance and independence among the protected attributes, the variance of the demographic parity $D_i$ is equal to the sum of the variances of $Y_i$ and $Y_{\bar{i}}$:
$\var(D_i) = \var(Y_i) + \var(Y_{\bar{i}}).$

\paragraph{No Collaboration.} In the absence of collaboration, agents prioritize their individual attributes
regardless of other agents' estimations.
More concretely, an agent $ i $ splits the input space into two distinct strata based on their binary attribute $X_i$.
These strata represent instances where the attribute is favored ($X_i = 1$) and unfavored ($X_i = 0$).
Within each stratum, homogeneity is assumed with respect to the characteristic of interest ($Y$), leading to the computation of average positive responses denoted as $Y_i$ or $Y_{\bar{i}}$.
Consequently, the variance of the \ac{DP} metric on attribute $ X_i $, denoted as $D_i$, in a setting without collaboration can be expressed as:
\begin{equation}\label{eq:varSelfish}
    \var(\hat{D}_i)_{\selfish} = \frac{1}{R_i}\sum_{j=1}^{R_i} (x_j - \hat{Y_i})^2 + \frac{1}{R_{\bar{i}}}\sum_{j=1}^{R_{\bar{i}}} (x_j - \hat{Y}_{\bar{i}})^2.
\end{equation}
Each agent's ability to measure $D_i$ is constrained by the choice of sample sizes $R_i$ and $R_{\bar{i}}$.
The used sampling strategies determines the values of $ R_i$ and $ R_{\bar{i}} $. All expressions of the total budget $R_i$ and $R_{\bar{i}}$ for different collaboration strategies and sampling methods are summarized in~\Cref{tab:samplingRi} in~\Cref{app:sampling_methods}.
Under uniform sampling, $R_i$ is typically set equal to $p_iR$, representing the probability of observing $X_i = 1$.
Conversely, in stratified sampling, a uniform allocation strategy assigns $R_i = R/2$, ensuring an equal budget allocation across all strata.
However, Neyman sampling introduces a more nuanced approach, wherein $R_i$ and $R_{\bar{i}}$ are strategically determined to minimize the expression defined in Eq.~\eqref{eq:varSelfish}.
Depending on the extent to which the division into two strata is relevant, Neyman sampling can, for example, give a distribution close to uniform, close to disproportionate sampling or something in-between~\cite{mathew2013efficiency}.

\textbf{\textit{\aposteriori} Collaboration.}
Under \aposteriori collaboration, agents face a similar situation regarding the variance of $ D_i $ as when not collaborating.

{\footnotesize
\begin{equation}\label{eq:VarApo}
    \var(\hat{D}_i)_{\aposteriori} =  \frac{1}{R_i}\sum_{j=1}^{R_i} (x_j - \hat{Y_i})^2 + \frac{1}{R_{\bar{i}}}\sum_{j=1}^{R_{\bar{i}}} (x_j - \hat{Y}_{\bar{i}})^2.
\end{equation}
}
 
We note that $R_i$ and $R_{\bar{i}}$ are expressed differently for the non-collaborative and \aposteriori collaborative settings.
The total budget $R_i$ allocated to each stratum is the sum of the budget allocated by the agent $i$ on the stratum and the total budget allocated by the other agents on the stratum. Thus, as each agent homogeneously considers its own attribute in its strata, an agent $j \neq i$ uniformly samples the attribute $X_i$. The budget on stratum $X_i = 1$ is $p_i(m-1)R$ added to the budget allocated by agent $i$ on this specific stratum.

In the case of stratified sampling, agent $i$ spends half the budget ($R/2$) on each stratum. In total, $R_i$ is thus equal to $R/2 +p_i(m-1)R$. Similarly, with Neyman sampling, the agent $i$ spends the optimal budget $R_i^*$ (obtained from Eq.~\eqref{eq:VarApo}) on stratum $X_i = 1$. In total, $R_i = R_i^* + p_i(m-1)R$. While for uniform sampling, agent $i$ samples the strata uniformly as the other agents, resulting in $R_i = p_imR$.

\paragraph{\textit{\apriori} Collaboration.}
For \apriori collaboration with \unif sampling, $ \var(\hat{D}_i) $ is the same as for \aposteriori collaboration with \unif sampling, as these situations are equivalent.
We now analyze \apriori collaboration with \stratified or \neyman sampling.
With \apriori collaboration, each agent considers the $n=2^m$ strata, encompassing all possible combinations of the protected attributes.
Unlike in \aposteriori collaboration, where agents treat their strata as homogeneous and uniformly sample them, in this scenario, $Y_i$ and $Y_{\bar{i}}$ are stratified, resulting in variances distinct from those observed in previous situations.
Specifically, we can express this variance as follows: 
{\small
\begin{equation}\label{eq:varApriori}
    \var(\hat{D}_i)_{\apriori} = \sum_{j=1}^n p_j^2 \left( \frac{1}{R_j} - \frac{1}{p_jN} \right) \var(\hat{D}_j)^2.
\end{equation}
}
For \stratified sampling, the budget on each stratum is $R_j = B/2^m$ because all agents spend an equal amount of requests on each stratum.
For \neyman sampling, we have $R_j = R_j^*$ with 
$\left(R_1^*, \dots, R_n^*\right) = argmin~\left(\var(\mu)_{\apriori}\right)$.

As the right-hand side of Eq.~\ref{eq:varApriori} does not depend on $i$, the variance $\var(\hat{D}_i)_{\apriori}$ is the same for all agents $i$.

Having established variances for each agent auditing its own attribute, we are now ready to compute the aggregate \ac{DP} variance across all agents of the auditor.
We recall that our overarching goal is to minimize the average variance in practical applications (see Section~\ref{s:background}).
To globally assess the interest of collaboration, we rely on the average \ac{DP} variance realized by agents:
\begin{definition}{Average DP variance.} According to Section~\ref{s:background}, for a set $I$ of collaborative agents where each agent
$(A_j)$ audits a demographic parity $D_i$ with variance $\var(\hat{D}_i)$, the average one is:
{\footnotesize
    \begin{equation}\label{eq:to_minimize}
        \var(\hat{D}) = \frac{1}{m}\sum_{i=1}^m \var(\hat{D}_i).
    \end{equation}
  }  
\end{definition}

\section{The Dynamics of Collaboration}\label{s:th-B}
We now outline the foundational principles behind the interplay between collaboration strategies and sampling methods.
Building upon our previous derivations, our main result comprises three theoretical outcomes that provide guidelines for collaborative fairness audits.

In the following, $\sum_j (x_j - \hat{Y}_i)^2$ will be denoted as $\sigma_i^2$ for brevity as each sample $x_j$ has an equal variance. This shorthand simplifies expressions for clarity in mathematical formulas.

\subsection{When Collaboration is Advantageous}
We introduce two theorems that describe when collaboration leads to a more accurate audit accuracy.

\begin{restatable}[]{theorem}{thone}\label{th1}
    \textbf{Except for \stratified sampling under \apriori collaboration, \aposteriori and \apriori collaboration leads to more accurate results}. Apart from one situation (see Theorem~\ref{th3}), collaboration is always beneficial and is an effective approach to increase the accuracy of fairness audits, \textit{i.e.} $Var(\hat{D})_{collab} \leq Var(\hat{D})_{\selfish}$.
\end{restatable}

Below we provide the proof of this result for the \aposteriori collaboration with \stratified sampling.

\paragraph{Proof.} 
As seen in Section~\ref{s:se-G}, the variance of the average $DP$ estimation $ \var(\hat{D})_{\aposteriori} $ in this setting can be written as:

{\scriptsize
$$\var(\hat{D})_{\aposteriori} = \frac{1}{m} \sum_{i=1}^m \Bigg(\frac{\sigma_i^2}{\frac{R}{2} + p_i(m-1)R} 
 + \frac{\sigma_{\bar{i}}^2}{\frac{R}{2}+p_{\bar{i}}(m-1)R}\Bigg).$$
}

Since $\forall i \in I, (m-1)p_iR > 0$ and $(m-1)p_{\bar{i}}R > 0$, the previous equation leads to the following inequality: 
\begin{align*}
\var(\hat{D})_{\aposteriori} &\leq \underbrace{\frac{1}{m} \sum_{i=1}^m \Bigg(\frac{\sigma_i^2}{\frac{R}{2}} 
 + \frac{\sigma_{\bar{i}}^2}{\frac{R}{2}}\Bigg)}_{:= \var(\hat{D})_{\selfish}}.
\end{align*}
Combining Eq.~\ref{eq:VarApo} and Eq.~\ref{eq:to_minimize}, the right-hand side of the above inequality is exactly the definition of the variance of $\hat{D}$ without collaboration and \stratified sampling (Eq.~\ref{eq:varSelfish}).

Thus, we have just proven that \aposteriori collaboration with stratified sampling is \emph{always} beneficial.
In \Cref{a:proofs}, we prove that all combinations of collaboration strategies and sampling methods (with the only exception of \apriori collaboration with stratified sampling) are beneficial over no collaborations.
We also show that for all collaborative strategies with \unif sampling, the variance on $\hat{D}$ linearly decreases with the number of collaborating agents. 

This linear reduction is also similarly observed for \apriori collaboration with \neyman sampling (Appendix~\ref{a:api:neyman}).

\paragraph{Conclusions.} Except in the case of \apriori collaboration with \stratified sampling, the gains from collaboration increases with the number of collaborating agents. 
The gains from collaboration can even be linear on $m$. 
It is therefore recommended that agents collaborate.

\begin{restatable}[]{theorem}{thtwo}\label{th2}
    \textbf{Under \aposteriori collaboration, \stratified and \neyman sampling methods are asymptotically equivalent to uniform sampling}. 
    The advantages of more advanced sampling methods vanishes with the increasing number of agents under \aposteriori collaboration:
    $\var(\hat{D})_{\stratified} \underset{m \rightarrow +\infty}{\sim}\var(\hat{D})_{\unif}$\\ and $\var(\hat{D})_{\neyman} \underset{m \rightarrow +\infty}{\sim} \var(\hat{D})_{\unif}$.
\end{restatable}

\paragraph{Proof.} We consider \aposteriori collaboration.
Under \stratified sampling, the agent $i$ splits equally her budget on the two strata: $R/2$ for $X_i = 1$ and $R/2$ for $X_i = 0$. 
This distribution does not depend on $m$. 
The total budget on these strata with \aposteriori collaboration with $m$ agents is $R_i = R/2 + (m-1)p_iR$. If $m \rightarrow +\infty$ then $R_i \sim mp_iR$ (and the same thing replacing $i$ by $\bar{i}$). Thus:

{\footnotesize
\begin{align*}
\var(\hat{D})_{\stratified} &= \frac{1}{m} \sum_{i=1}^m \Bigg(\frac{\sigma_i^2}{\frac{R}{2} + p_i(m-1)R} + \frac{\sigma_{\bar{i}}^2}{\frac{R}{2}+p_{\bar{i}}(m-1)R}\Bigg)\\
 &\underset{m \rightarrow +\infty}{\sim} \underbrace{\frac{1}{m} \sum_{i=1}^m \Bigg(\frac{\sigma_i^2}{p_imR}+ \frac{\sigma_{\bar{i}}^2}{p_{\bar{i}}mR}\Bigg)}_{:= \var(\hat{D})_{\unif}},
\end{align*}
}
which is exactly $\var(\hat{D})_{\aposteriori}$ with \unif sampling. Therefore, when using \aposteriori collaboration with a large number of agents, each agent can simply adopt uniform sampling for its queries. 
The proof for \aposteriori collaboration with \neyman sampling follows similarly, where $R/2$ is replaced by $R_i^*$ (\Cref{a:proofs2}). 

\paragraph{Conclusions.} We know that without missing knowledge on $\mathcal{A}$, the best sampling method is \neyman sampling. As the variance with \neyman sampling converges to that of \unif, the benefits of extra-information on $\mathcal{A}$ vanishes with the increasing number of agents under \aposteriori collaboration. Thus, if the number of agents collaborating is large in \aposteriori collaboration, each agent can simply do \unif sampling.

\subsection{When Collaboration is Disadvantageous}
We now highlight potential issues in the collaboration.
This section focuses on \apriori collaboration with \stratified sampling, which intuitively seems like a desirable candidate in practical settings.
This is because of its coordinated nature, with the fairest sampling method that is compatible with black-box assumptions.

The forthcoming theorem leverages the following observation:
\paragraph{Observation 1.} \textit{\textbf{For any collaboration with $m$ agents, there is a stratum among the existing $2^m$ that represents at least $1/(2m)$ of the population.}
It is mathematically expressed as: $$\exists j^*, 1 \leq j^* \leq 2^m, p_{j^*} \geq \frac{1}{2m}.$$
}

\begin{figure}[t]
	\centering
		\includegraphics[height = 4cm,width=\linewidth]{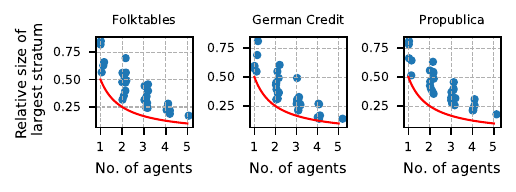}
   
 \caption{\centering The relative size of the largest stratum for all possible $m$ auditor configurations and three datasets. The red curve is $y=\frac{1}{2x}$.}
	\label{fig:obs1}
\vspace{0.44cm}
\end{figure}

This observation formalizes that in collaborations involving $m$ agents, there will typically be certain subgroups or strata that represent a significant portion of the overall population. As the number of agents $m$ increases, the number of potential collaboration configurations grows exponentially, reaching $2^m$. Yet, as $ m $ increases, strata become unbalanced with some stratum being consistently larger than the average $2^{-m}$ fraction of the dataset.
We verify this also within the three classical datasets used for our experiments with binarized attributes in~\Cref{s:exp}, and even with non-binary attributes (\Cref{appendix:add_exps_multi_class}).
Figure~\ref{fig:obs1} shows the relative size of a largest stratum for each possible assignment of $m$ agents to the five protected attributes considered in each dataset.
Any point below the red line would violate Observation~1.
Hence, all of the 
$\sum_{m=1}^5\binom{5}{m}=31$ configurations tested for each dataset confirm Observation 1.

\begin{theorem}\label{th3}
    \textbf{The \apriori collaboration can be disadvantageous.} The variance on the estimator using \stratified \apriori increases with the number of agents.
    
    If $\forall m > 0, \exists j^*, 1 \leq j^* \leq 2^m, p_{j^*} \geq \frac{1}{2m},$
    (i.e. Observation 1 holds) then $Var(\hat{D})_{\apriori}^{\stratified}\underset{m \rightarrow \infty}{\rightarrow} +\infty $.
\end{theorem}
\textbf{Proof.} With \apriori collaboration and \stratified sampling, the variance of $DP$ is $\var(\hat{D}) = \frac{1}{m} \sum_{j=1}^{2^m} p_j^2 \left(\frac{2^m}{B} - \frac{1}{p_jN}\right)Var(\hat{D}_j)^2$.

It corresponds to Eq.~\eqref{eq:varApriori} for $R_i = B/2^m$ summed of all agents. As we are interested in show that the variance of the estimator increases, we lower bound it:
$\var(\hat{D}) > \frac{1}{m} \sum_{j=1}^{2^m} p_j^2 \left(\frac{2^m}{B}-1\right)Var(\hat{D}_j)^2$.
In particular, under Observation 1, there is a stratum among the existing $2^m$ that represents at least $1/(2m)$ of the population. That means that $\exists j^*, 1 \leq i \leq 2^m, p_j \geq \frac{1}{2m}$. The sum of the variances is at least greater than the variance of each of its components: $\var(\hat{D}) > \frac{2^{m-2}}{Bm^3} Var(\hat{D}_{j^*})^2$.
In the majority of cases, the decisions of $\mathcal{A}$ on this stratum are not always the same, so $Var(\hat{D}_{j^*}) \neq 0$. Thus, $\frac{2^{m-2}}{Bm^3} Var(\hat{D}_{j^*})^2 \underset{m \rightarrow \infty}{\rightarrow} + \infty$ and so does $\var(\hat{D})$.

\paragraph{Conclusions.}
Under Observation 1, Theorem~\ref{th3} demonstrates that the estimator variance tends to go to infinity with \apriori collaboration and \stratified sampling.
Although this may appear as an appealing strategy, it turns out to be detrimental to the general audit accuracy in realistic settings where strata are severely unbalanced.
More precisely, the strategy of allocating a constant number of samples per stratum turns out detrimental as $m$ grows since the number of samples allocated to the majority stratum decreases exponentially with $m$ whereas its size (and hence its contribution to the overall estimation variance) decreases only proportionally to $m$. While our theoretical results characterizes the asymptotic behaviour as $m$ grows, our experimental results (see Section~\ref{s:exp}) already show this behaviour for low values of $m$.\\

Counter-intuitively, we thus find that \apriori collaboration is not the best strategy to consider.
Its variance with stratified sampling increases with the number of agents and its combination with \neyman sampling is impossible in practice.
Besides, \apriori collaboration with \unif sampling is equivalent to \aposteriori collaboration with \unif sampling by definition.
On the contrary, \aposteriori collaboration exhibits advantages with all sampling methods. 
We also showed that the advantages of advanced sampling methods vanish when the number of agents is large. 
So when agents collaborate, if there are many, they have every interest in using \aposteriori collaboration with \unif sampling.

\section{Experiments and Results}\label{s:exp}

\begin{figure}[b]
	\centering
		\includegraphics[width=.97\linewidth]{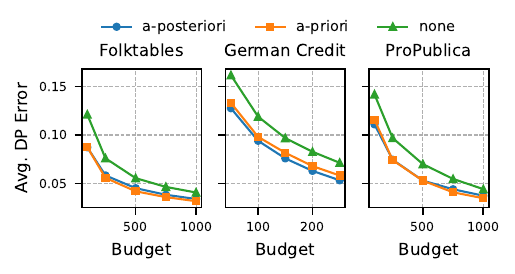}
   
  \caption{\centering
  2-agent collaboration with stratified sampling. The budget ranges are relative to the size of the dataset being studied.
  We observe that collaboration (\aposteriori and \apriori) can significantly improve \ac{DP} error. This is in line with \Cref{th1}.
 }
	\label{fig:thrm1}
\vspace{0.44cm}
\end{figure}

\begin{figure}[t]
	\centering
		\includegraphics[width=.97\linewidth]{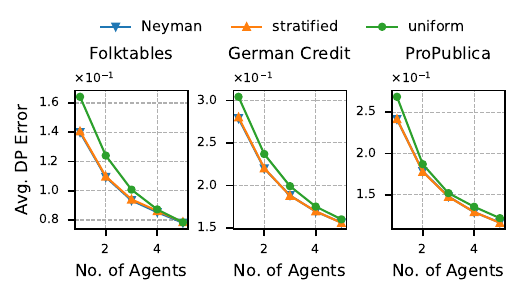}
   
 \caption{\centering
 Different sampling methods with \aposteriori collaboration.
 The more agents collaborate, the more all sampling methods tend to converge. This is in line with \Cref{th2}.
 }
	\label{fig:thrm2}
\vspace{0.44cm}
\end{figure}

To empirically evaluate and understand collaboration with real-world datasets, we implement a simulation framework.
We leverage three datasets: \gc~\cite{misc_statlog_(german_credit_data)_144}, \pp~\cite{angwin2016} and \ft~\cite{ding2021retiring} (full description is deferred to~\Cref{appendix:exp_setup}).
We consider five attributes on each dataset where each attribute is binary by default or binarized by following a certain scheme (see~\Cref{appendix:exp_setup}).
The labels for the prediction task in each dataset are also binary.
To simulate black-box models, we adopt a unique strategy by treating dataset labels as responses from the \ac{ML} model, avoiding the traditional need to train specific models for each audit task. 
This approach views datasets as extensive passive sampling sets of the target model, eliminating the need to select a training algorithm and an ML model from diverse choices.
We open source all source code and documentation~\cite{deVos2024fairnessCode}.

\textbf{Setup.}
We consider combinations of the three sampling methods described in~\Cref{ss:sampling} and the  collaboration strategies described in~\Cref{s:se-G}.
The \ft dataset is composed of \num{5916565} samples, \gc of \num{1000} and \pp of \num{6172} samples.
We run each experiment for \num{300} repetitions.
These repetitions yield a good balance between accuracy and computational efficiency.
Our experiments report the average \ac{DP} error, whose minimization is the objective of an auditor (see~\Cref{s:background}).

\subsection{Impact of Collaboration with Two Agents}
\label{subsec:exp_thrm1}
We simulate different collaboration strategies with stratified sampling for all three datasets and observe the average \ac{DP} error for different per-agent query budgets $ R $.
The query budgets are varied depending on the dataset, ranging from \num{100} to \num{1000} for the \ft and the \pp while for the \gc dataset, we vary from \num{50} to \num{250} given its small size. 

For this experiment, two agents audit a particular attribute in each dataset; we consider gender and marital status for the \ft dataset, age and gender for the \gc dataset and lastly, gender and African-American origin for the \pp dataset.
These results are shown in~\Cref{fig:thrm1} where each column represents a different dataset.
Our first observation is that for all configurations, the average \ac{DP} error decreases as $ R $ increases since individual \ac{DP} estimations become closer to the ground-truth value. 
Secondly, we observe that \aposteriori or \apriori collaboration always decreases the average \ac{DP} error compared to when not collaborating and therefore \emph{it is always beneficial to collaborate}.

This behaviour is consistent across all three datasets, providing strong empirical evidence for \Cref{th1}. 
The average \ac{DP} error reduces anywhere between $17.4\%$ to $24.6\%$ by collaboration.
Furthermore, we observe that both the collaborative strategies have similar performance in this two-agent setting. 
We further extend this comparison in \Cref{subsec:exp_thrm3} while considering more agents.

\subsection{Performance of Different Sampling Methods with Multi-agent Collaboration}
\label{subsec:exp_thrm2}
This experiment aims to observe how the average \ac{DP} error changes for different sampling methods as we increase the level of collaboration.
We consider the uncoordinated collaboration \ie the \aposteriori strategy in this section and analyse the coordinated collaboration \ie \apriori strategy in the following section.
For each dataset, we vary the number of collaborating agents ($m$) from \num{1} (no collaboration) to \num{5}.
For each specific value of $m$, we report the average over all $\binom{5}{m}$ possible combinations of $m$ collaborating agents. For example, there are $10$ combinations of protected attributes when $m=2$.
Each combination is still run several times with different random seeds, as previously described.
We set the budget $R = 500$ for the \ft dataset, $R = 100$ for the \gc dataset and $R = 250$ for the \pp dataset.

\begin{figure}[t]
	\centering
		\includegraphics[width=.98\linewidth]{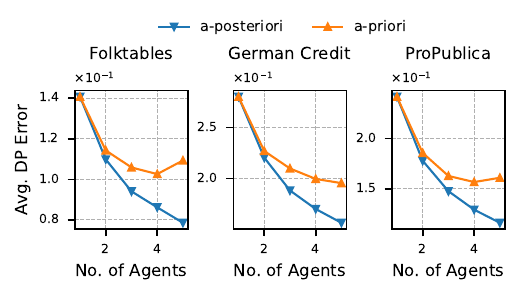}
   
 \caption{\centering
 Different collaborative strategies with stratified sampling.
We observe that as more agents collaborate, the \apriori strategy can be disadvantageous. This is in line with \Cref{th3}.
 }
\label{fig:thrm3}
\vspace{0.44cm}
\end{figure}

\Cref{fig:thrm2} displays the results of all sampling methods across different datasets, with a column per dataset. 
We note that the average \ac{DP} error for all methods decreases as collaboration increases, reinforcing \Cref{th1}. 
Stratified and Neyman sampling consistently shows lower error rates than uniform sampling when $m$ is small, highlighting the advantages of stratification.
However, as $m$ increases, these advantages diminish, and the performance gap with uniform sampling methods narrows significantly. 
This convergence in performance, anticipated in \Cref{th2} as $m \rightarrow +\infty$, is observed empirically even at moderate values of $m$. 
For example, at $m = 5$, the performance of uniform and stratified sampling is alike for the \ft dataset and closely matched for the other datasets. 
This empirical evidence confirms \Cref{th2}.
Lastly, we also note that while Neyman sampling performs optimally, the empirical difference of average \ac{DP} error with stratified sampling is very low $(< 0.001)$.

\subsection{Performance of different collaborative strategies with multi-agent collaboration}
\label{subsec:exp_thrm3}

In this section we thoroughly examine the coordinated collaboration strategy \ie \apriori strategy in comparison to the uncoordinated collaboration strategy \ie \aposteriori.
We keep the setup same as \Cref{subsec:exp_thrm2} and observe the average \ac{DP} error when increasing the number of collaborating agents from $ m = 1$ to $m = 5$. 
\Cref{fig:thrm3} depicts our results for stratified sampling.
We include the results with Neyman sampling in \Cref{fig:apr_apo_neyman} (\Cref{appendix:add_exps}).
We observe that, under stratification, the error of \apriori shows a decreasing trend as $m$ grows from \num{2} to \num{4} but then either reduces very little or exhibits an increasing trend at full collaboration $(m = 5)$.
This is particularly evident for the \ft dataset.
While \Cref{th3} shows that the estimator variance tends to infinity as $m \rightarrow + \infty$, we observe the detrimental effects of \apriori collaboration with stratified sampling even for $m = 5$ agents.
On the other hand, the error of \aposteriori under stratified sampling always decreases with increasing collaboration.
Thus contrary to expectations, \textit{extensive coordination using the \apriori approach can be disadvantageous}, whereas simpler, uncoordinated collaboration consistently proves beneficial.

\section{Conclusions and Future Work}
Decision-making algorithms and models are now widespread online and often lack transparency in their operation. 
Multiple regulatory bodies are willing to conduct efficient fairness audits. 
However, agents can only estimate fairness attributes since they have a hard cap on the number of queries they can issue.
This paper shows that collaboration among previously independent audit tasks can yield a substantial gain in accuracy under fixed query budgets.
We observed and analyzed an interesting case where prior agent coordination on the queries causes worse outcomes than a non-coordinated collaboration strategy.
We also show that, in practice, the latter strategy performs nearly as well as the (infeasible) optimal strategy, underlining the relevance of that proposed collaboration strategy.

Future work directions include exploring the intersectional fairness and review \apriori collaboration in this context~\cite{10.24963/ijcai.2023/742}. 
Additionally, collaboration using active approaches, like adaptive sampling, could yield efficient and accurate audits. However, this may entail higher synchronization costs.

\clearpage
\bibliography{ref}

\clearpage
\onecolumn
\appendix

\clearpage
\onecolumn

\section{Sampling Methods}
\label{app:sampling_methods}
In~\Cref{s:se-G}, the variance formulas have been driven of all sampling methods (\unif, \stratified and \neyman) with all the collaboration strategies (\selfish, \aposteriori and \apriori). In particular, Equations~\ref{eq:varSelfish} \ref{eq:VarApo} and~\ref{eq:varApriori} depend on the number of queries on each subpopulation. The following table sums up the total queries allocation $R_i$ for each subpopulation $X_i = 1$ with $m$ agents. All budget $R_{\bar{i}}$ are the complementary proportions of the total budget ($R$ for \selfish, $B$ for \aposteriori and \apriori collaboration).

\begin{table}[h!]
\centering
\caption{\centering Number of queries $R_i$ in the subpopulation $X_i = 1$ (left table) and $R_{\bar{i}}$ in the subpopulation $X_i = 0$ (right table). The notation $R_i^*$ stands for $R_i^* = argmin~\left(\var(\hat{DP}_i)\right)$.}
    \label{tab:samplingRi}
    
\begin{tabular}{|c| ccc|}
                \hline
                \multicolumn{4}{|c|}{$R_i$}\\
                    \hline
                   & \textbf{\unif} & \textbf{\stratified} & \textbf{\neyman} \\
                  \hline
                  \selfish & $P_iR$ & $R/2$ & $R_i^*$\\
                  \aposteriori & $mP_iR$ & $R/2 + (m-1)P_iR$ & $R_i^* + (m-1)P_iR$\\
                  \apriori & $mP_iR$  & $Rm/2$ &  $m~R_i^*$ \\
                  \hline
\end{tabular}
\quad
\begin{tabular}{|c|c|}
\hline
& $R_{\bar{i}}$\\
\hline
\selfish & $R - R_i$\\
\aposteriori & $B - R_i$\\
\apriori & $B - R_i$\\
\hline
\end{tabular}
\end{table}
\section{Proofs of~\Cref{th1}}\label{a:proofs}
This section provides the proof of~\Cref{th1}(in~\Cref{s:th-B}) and some complementary results in all cases. 
\thone*

\subsection{\aposteriori collaboration}\label{a:th1:apo}
\textbf{Proof.}  As seen in~\Cref{s:se-G}, the variance of the average $DP$ estimation $ \var(\hat{D})$ with any sampling method and collaboration can be written as:

\begin{align*}
\var(\hat{D}) &= \frac{1}{m} \sum_{i=1}^m \Bigg(\frac{\sigma_i^2}{R_i} 
 + \frac{\sigma_{\bar{i}}^2}{R_{\bar{i}}}\Bigg)
\end{align*}
We refer to \Cref{tab:samplingRi} for the budget expressions $R_i$ depending on the sampling methods and collaboration.

\subsubsection{\unif sampling}\label{a:apo:unif}
With \unif sampling, $R_i = mp_iR$ and $R_{\bar{i}} = m(1-p_i)R$. So

\begin{align*}
\var(\hat{D})_{\aposteriori}^{\unif} &= \frac{1}{m} \sum_{i=1}^m \Bigg(\frac{\sigma_i^2}{mp_iR} 
 + \frac{\sigma_{\bar{i}}^2}{m(1-p_i)R}\Bigg)
\end{align*}
By factoring by $1/m$ we find $\var(\hat{D})_{\selfish}^{\unif}$:
\begin{align*}
\var(\hat{D})_{\aposteriori}^{\unif} &= \frac{1}{m}\Bigg(\underbrace{\frac{1}{m} \sum_{i=1}^m \Bigg(\frac{\sigma_i^2}{p_iR} 
 + \frac{\sigma_{\bar{i}}^2}{(1-p_i)R}\Bigg)}_{:= \var(\hat{D})_{\selfish}^{\unif}}\Bigg)
\end{align*}

Thus ($m > 1$), $\var(\hat{D})_{\aposteriori}^{\unif} < \var(\hat{D})_{\selfish}^{\unif}$. We even demonstrate that using the \unif sampling, the variance of \aposteriori collaboration decreases by a factor $m$ compared to \selfish.

\subsubsection{\stratified sampling}

\textit{It is the same proof as in~\Cref{s:th-B}.}

\par With \stratified sampling, $R_i = \frac{R}{2} + p_i(m-1)R$ and $R_{\bar{i}} = \frac{R}{2}+p_{\bar{i}}(m-1)R$. So

\begin{align*}
\var(\hat{D})_{\aposteriori}^{\stratified} &= \frac{1}{m} \sum_{i=1}^m \Bigg(\frac{\sigma_i^2}{\frac{R}{2} + p_i(m-1)R} 
 + \frac{\sigma_{\bar{i}}^2}{\frac{R}{2}+p_{\bar{i}}(m-1)R}\Bigg)
\end{align*}

Since $\forall i \in I, (m-1)p_iR > 0$ and $(m-1)p_{\bar{i}}R > 0$, the previous equation leads to the following inequality:
\begin{align*}
\var(\hat{D})_{\aposteriori}^{\stratified} &< \underbrace{\frac{1}{m} \sum_{i=1}^m \Bigg(\frac{\sigma_i^2}{\frac{R}{2}} 
 + \frac{\sigma_{\bar{i}}^2}{\frac{R}{2}}\Bigg)}_{:= \var(\hat{D})_{\selfish}^{\stratified}}
\end{align*}
Combining~\Cref{eq:VarApo} and~\Cref{eq:to_minimize}, the right-hand side of the above inequality is exactly the definition of $\var(\hat{D})_{\selfish}^{\stratified}$, or the variance of the average $DP$ without collaboration with \stratified sampling.

\subsubsection{\neyman sampling}
\par With \neyman sampling, $R_i = R_i^* + (m-1)p_iR$ and $R_{\bar{i}} = (R - R_i^*) + (m-1)p_iR$. We note The proof is identical to that of \stratified sampling.

The variance of the average $DP$ estimation $ \var(\hat{D})_{\aposteriori}^{\neyman} $ can be written as:

\begin{align*}
\var(\hat{D})_{\aposteriori}^{\neyman} &= \frac{1}{m} \sum_{i=1}^m \Bigg(\frac{\sigma_i^2}{R_i^* + (m-1)p_iR} 
 + \frac{\sigma_{\bar{i}}^2}{(R - R_i^*) + (m-1)p_iR}\Bigg)
\end{align*}

Since $\forall i \in I, (m-1)p_iR > 0$ and $(m-1)p_{\bar{i}}R > 0$, the previous equation leads to the following inequality:
\begin{align*}
\var(\hat{D})_{\aposteriori}^{\neyman} &< \underbrace{\frac{1}{m} \sum_{i=1}^m \Bigg(\frac{\sigma_i^2}{R_i^*} 
 + \frac{\sigma_{\bar{i}}^2}{R - R_i^*}\Bigg)}_{:= \var(\hat{D})_{\selfish}^{\neyman}}
\end{align*}
Combining~\Cref{eq:VarApo} and~\Cref{eq:to_minimize}, the right-hand side of the above inequality is exactly the definition of $\var(\hat{D})_{\selfish}$, or the variance of the average $DP$ without collaboration with \stratified sampling.

\subsubsection{Conclusion}
Thus, \aposteriori collaboration is \emph{always} beneficial whether with \unif, \stratified or \neyman. We even demonstrate that using the \unif sampling, the variance of \aposteriori collaboration decreases by a factor $m$ compared to \selfish.

\subsection{\apriori collaboration}\label{a:th1:api}
Let us now move on to the case of \apriori collaboration. We treat the \unif sampling and \neyman sampling. The case of \stratified sampling is a specific case leading to~\Cref{th3} which we will prove in~\Cref{a:th1:apo}.

\subsubsection{\unif sampling}\label{a:api:unif}
As established in \Cref{ss:sampling-variance}, the sampling variance of \apriori collaboration with \unif sampling, $\var(\hat{D})_{\apriori}^{\unif}$ is the same as for \aposteriori collaboration with
\unif sampling, $\var(\hat{D})_{\aposteriori}^{\unif}$, as these situations are equivalent. The result proved in \Cref{a:apo:unif} shows that $\var(\hat{D})_{\apriori}^{\unif} < \var(\hat{D})_{\selfish}^{\unif}$. We even demonstrate that using the \unif sampling, the variance of \apriori collaboration decreases by a factor $m$ compared to \selfish.

\subsubsection{\stratified sampling}

It is the specific case leading to~\Cref{th3}.

\subsubsection{\neyman sampling}\label{a:api:neyman}

The sampling variance with \apriori collaboration is defined in \Cref{eq:varApriori} as:
$$\var(\hat{D}_i)_{\apriori} = \sum_{j=1}^n p_j^2 \left( \frac{1}{R_j} - \frac{1}{p_jN} \right) \sigma_j^2$$

\par With \neyman sampling, $R_j = mR_j^*$. The variance of the average $DP$ estimation $ \var(\hat{D})_{\apriori}^{\neyman} $ can be written as:

\begin{align*}
\var(\hat{D}_i)_{\apriori}^{\neyman} &= \sum_{j=1}^{2^m} p_j^2 \left( \frac{1}{mR_j^*} - \frac{1}{p_jN} \right) \sigma_j^2\\
    &\leq \frac{1}{m}\sum_{j=1}^{2^m} \frac{\sigma_j^2}{R_j^*}
\end{align*}
The inequality is obtained by harsh bounds ($\forall j, p_j^2 \leq 1$ and $\frac{1}{p_jN} > 0$). It means that the sampling variance of \apriori collaboration with \neyman sampling is lower than the unweighted sum of the sampling variance of the $2^m$ stratas.

In the other hand, the sampling variance of \selfish with \neyman sampling is: $\var(\hat{D})_{\selfish} = \frac{1}{m}\sum_{i=1}^m \left(\frac{\sigma_i^2}{R_i^*} + \frac{\sigma_{\bar{i}}^2}{R - R_i^*} \right)$ (\Cref{eq:varSelfish} in \Cref{eq:to_minimize} with $R_i$ and $R_{\bar{i}}$ defined with \Cref{tab:samplingRi}). The sum can be splits on the $2^m$ strata considered in the collaboration (the intersection of all possible subpopulation). In that case, all strata are counted once per agent so $m$ times in the global sum: $\var(\hat{D})_{\selfish} = \frac{1}{m}\sum_{j=1}^{2^m} m\frac{\sigma_j^2}{R_j^*}$. We can thus write $\var(\hat{D})_{\selfish} = m\var(\hat{D}_i)_{\apriori}^{\neyman}$ \textit{i.e.} $\var(\hat{D}_i)_{\apriori}^{\neyman} \leq \frac{1}{m} \var(\hat{D})_{\selfish}$.

We even demonstrate that using the \neyman sampling, the variance of \apriori collaboration decreases at least by a factor $m$ compared to \selfish.

\subsubsection{Conclusion}
\apriori collaboration is beneficial with \unif or \neyman. We even demonstrate that for those two sampling methods, the variance of \apriori collaboration decreases by a factor $m$ compared to \selfish.

\section{Proofs of~\Cref{th2}}\label{a:proofs2}

\thtwo*

This theorem has been proven for \aposteriori collaboration with \stratified sampling in~\Cref{s:th-B}. The proof is exactly the same for $\neyman$ sampling with \aposteriori collaboration by replacing $R/2$ by $R_i^*$ in the proof:

\textbf{Proof.} We consider \aposteriori collaboration.
Under \neyman sampling, agent $i$ splits her budget on the two strata as follows: $R_i^*$ for $X_i = 1$ and $B - R_i^*$ for $X_i = 0$. This distribution does not depend on $m$ The total budget on these strata, with \aposteriori collaboration with $m$ agents is $R_i = R_i^* + (m-1)p_iR$. If $m \rightarrow +\infty$ then $R_i \sim mp_iR$ (and the same thing replacing $i$ by $\bar{i}$). Thus:
\begin{align*}
\var(\hat{D})_{\aposteriori}^{\neyman} &= \frac{1}{m} \sum_{i=1}^m \Bigg(\frac{\sigma_i^2}{R_i^* + p_i(m-1)R} + \frac{\sigma_{\bar{i}}^2}{R-R_i^*+p_{\bar{i}}(m-1)R}\Bigg)\\
 &\underset{m \rightarrow +\infty}{\sim} \underbrace{\frac{1}{m} \sum_{i=1}^m \Bigg(\frac{\sigma_i^2}{p_imR}+ \frac{\sigma_{\bar{i}}^2}{p_{\bar{i}}mR}\Bigg)}_{:= \var(\hat{D})_{\aposteriori}^{\unif}}
\end{align*}
which is exactly $\var(\hat{D})_{\aposteriori}^{\unif}$.
Therefore, when using \aposteriori collaboration and the number of agents is large, each agent can adopt uniform sampling for its queries.

\newpage

\section{Additional Notes on Experiment Setup}
\label{appendix:exp_setup}

\subsection{Description of the leveraged datasets}
We conduct our study using three datasets: \gc~\cite{misc_statlog_(german_credit_data)_144}, \pp~\cite{angwin2016} and \ft~\cite{ding2021retiring}.
In the \gc dataset, the task involves predicting the creditworthiness of loan applicants.
Within the \pp dataset, we consider the recidivism risk task, predicting whether an individual will re-offend within two years after their initial criminal involvement.
In the \ft dataset, we consider the ACSPublicCoverage task that predicts whether low-income individuals, ineligible for Medicare, are covered by public health insurance. 
Attributes, such as age, gender, and demographic information, among others, are employed in these prediction tasks.
While some attributes are inherently binary, we binarize others by grouping values.
For instance, the marital status attribute in the \ft dataset is binarized as \num{1} for married and \num{0} for other statuses (widowed, divorced, separated, or never married).
In total, after binarization, we have five attributes corresponding to five auditing agents for each dataset.
Lastly, the prediction labels for each of the above tasks are also binary.
A comprehensive summary of the adopted datasets and the binarization strategy for each attribute can be found in \Cref{appendix:ft,appendix:gc,appendix:pp}.

In practice, the agents audit a platform through a black-box model.
To simulate such an audit, we must train a model for each task to be audited later.
In this work, we take a different approach and consider the labels in the dataset to be the response of the \ac{ML} model when queried with the corresponding attributes.
In other words, the datasets can be interpreted as a large passive sampling set of the target model (in our case, the real process that generated a given dataset). 
This strategy prevents the need of having to choose a training algorithm along with a ML model, among the diverse array of choices that exist.

\subsection{\ft dataset}
\label{appendix:ft}
In the \ft dataset~\cite{ding2021retiring}, we address the ACSPublicCoverage task, predicting whether a low-income individual without Medicare eligibility is covered by public health insurance.
We consider the following five attributes for auditing: gender, marital status, age, nativity and  mobility status. 
Their binarisation scheme is detailed in \Cref{tab:folktable_attr}.

\begin{table}[h]
    \caption{\centering Attributes in the \ft \task task. The value to description mapping for the original values can be found in~\cite{ding2021retiring}.}
    
    \centering
    \begin{tabular}{c|c|c}
    \toprule
    Attribute $X_i$ & How was it binarized ? & $P(X_i=1)$ \\
    \midrule
       \gender  & Binary by default & $0.43$ \\ 
       \nativity & Binary by default & $0.85$ \\
       \mig & Class $1$ is original value $\{1\}$ and $0$ for original values $\{\text{N/A}, 2, 3\}$ & $0.82$ \\
       \agep & Class $1$ is when $age \geq 25$ and $0$ when $age < 25$ & $0.66$ \\
       \mar & Class $1$ is original value $\{1\}$ and $0$ for original values $\{2, 3, 4, 5\}$ & $0.37$ \\
     \bottomrule
    \end{tabular}
    \label{tab:folktable_attr}
\end{table}

\subsection{\gc dataset}
\label{appendix:gc}
The task in the \gc dataset involves predicting whether a given individual is a good or bad credit risk~\cite{misc_statlog_(german_credit_data)_144}.
We chose the following five attributes for auditing: age, gender, marital status, whether the person has own telephone and employment status.
Their binarisation scheme is detailed in \Cref{tab:gc_attr}.

\begin{table}[h]
    \caption{\centering Attributes in the \gc dataset. More information regarding the dataset can be found in~\cite{misc_statlog_(german_credit_data)_144}.}

	\centering
	\begin{tabular}{c|c|c}
		\toprule
		Attribute $X_i$ & How was it binarized ? & $P(X_i=1)$ \\
		\midrule
		Own telephone  & Class is $0$ when original value is `none' and $1$ otherwise & $0.40$ \\ 
		Marital status & Class is $0$ for original value `single' and $1$ otherwise & $0.45$ \\
		Gender & Class is $0$ when original value is `female' and $1$ otherwise & $0.69$ \\
		Age & Class is $1$ when $age > 25$ and $0$ when $age \leq 25$ & $0.81$  \\
		Employment status & Class is $1$ for original values $\geq 4$ and $0$ otherwise  & $0.42$ \\
		\bottomrule
	\end{tabular}
	\label{tab:gc_attr}
\end{table}

\subsection{\pp dataset}
\label{appendix:pp}
The recidivism risk task in the ProPublica dataset involves predicting whether an individual will re-offend within 2 years after their initial criminal involvement~\cite{angwin2016}.
We consider the following five attributes for auditing: female, African-American origin, age below twenty five, misdemeanor and number of prior crimes.
Their binarisation scheme is detailed in \Cref{tab:pp_attr}.

\begin{table}[h]
    \caption{\centering Attributes in the \pp dataset. More information regarding the dataset can be found in~\cite{angwin2016}.}
	\centering
	\begin{tabular}{c|c|c}
		\toprule
		Attribute $X_i$ & How was it binarized ? & $P(X_i=1)$ \\
		\midrule
		Female & Binary by default & $0.19$ \\
		Misdemeanor & Binary by default & $0.36$ \\
		African-American  & Binary by default & $0.51$ \\
		Age below twenty five & Binary by default & $0.22$ \\
		Number of prior crimes & Class is $1$ if original value $> 0$ and $0$ otherwise  & $0.66$ \\
		\bottomrule
	\end{tabular}
	\label{tab:pp_attr}
\end{table}
\clearpage
\section{Additional Experiments}
\label{appendix:add_exps}

\subsection{On Neyman sampling}
In this section we include the results for \apriori and \aposteriori collaboration with Neyman sampling, expanding on results in \Cref{subsec:exp_thrm3}.
We observe that while \apriori strategy with stratified sampling can perform poorly (\Cref{fig:thrm3}), \apriori in combination with Neyman sampling always performs optimally as expected.
However, Neyman sampling is infeasible in practice as we discussed in \Cref{subsec:neyman_sampling}.

\begin{figure}[h]
	\centering
		\includegraphics[]{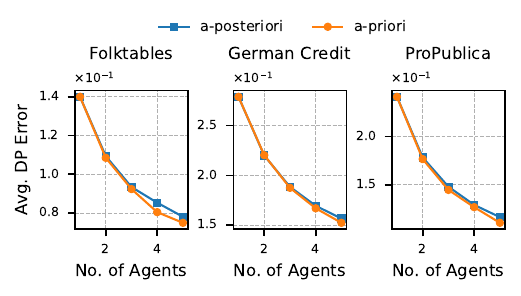}
   
 \caption{\centering
 Different collaborative strategies with Neyman sampling.
 We observe that \apriori strategy with Neyman sampling performs optimally as expected.
 However, it remains infeasible in practice.
 }
\label{fig:apr_apo_neyman}
\vspace{0.3cm}
\end{figure}

\subsection{On multi-valued attribute setting}
\label{appendix:add_exps_multi_class}

\begin{figure}[h!]
    \centering
    \includegraphics{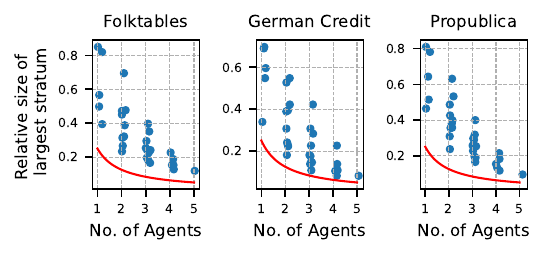}
    \caption{\centering The relative size of the largest stratum for non-binary attributes in the three datasets. The regression curve is $y = \frac{1}{4x}$.}
    \label{fig:multi_class}
\end{figure}

In this paper, we assume that agents audit only binary attributes. This assumption is not always verified in reality. As there is no clear consensus on how to define fairness for non-binary attributes, we considered demographic parity (DP) as the standard fairness metric and we binarized attributes in the experiments. However, to add some insight on non-binary attributes, we show in \Cref{fig:multi_class} that Observation~1 also holds in multi-valued attribute settings.

For example, rather than binarizing the `Employment status' attribute in the \gc dataset, we retain the five original groups as distinct values for this multi-valued attribute.
Similarly, the `Age' attribute in the German Credit dataset is divided into three groups: $\{< 25, [25-50], > 50\}$.
\Cref{tab:folktable_attr_2,tab:gc_attr_2,tab:pp_attr_3} present all the multi-valued attributes in the \ft, the \gc and the \pp datasets respectively.
In \Cref{fig:multi_class}, we observe that even in the multi-valued attribute case, the largest stratum represents a significant portion of the overall population.
Thus, \apriori collaborations may be disadvantageous in other fairness audits, even when the attributes are not binary.

\newpage
\begin{table}[h!]
    \caption{\centering Multi-valued attributes in the \ft \task task.}
    
    \centering
    \begin{tabular}{c|c}
    \toprule
    Attribute $X_i$ & Total classes and their relation to original attribute values  \\
    \midrule
       \gender  & Binary by default \\
       \nativity & Binary by default \\
       \mig & 3 classes corresponding to original values $\{1,2,3\}$ \\
       \agep & 3 classes corresponding to $age < 25$, $age \in [25, 50]$, $age > 50$ \\
       \mar & 5 classes corresponding to original values $\{1, 2, 3, 4, 5\}$ \\
     \bottomrule
    \end{tabular}
    \label{tab:folktable_attr_2}
\end{table}

\begin{table}[h!]
    \caption{\centering Multi-valued attributes in the \gc dataset.}

	\centering
	\begin{tabular}{c|c}
		\toprule
		Attribute $X_i$ & Total classes and their relation to original attribute values \\
		\midrule
		Own telephone  & Binary by default \\ 
		Marital status & 4 classes corresponding to $\{$`single', `div/dep/mar', `div/sep' and `mar/wid'$\}$ \\
		Gender & Binary by default  \\
		Age & 3 classes corresponding to $age < 25$, $age \in [25, 50]$, $age > 50$  \\
		Employment status & 5 classes corresponding to the bins $< 1, [1,4), [4,7),\geq 7$ and `unemployed' \\
		\bottomrule
	\end{tabular}
	\label{tab:gc_attr_2}
\end{table}

\begin{table}[h!]
    \caption{\centering Multi-valued attributes in the \pp dataset.}
	\centering
	\begin{tabular}{c|c}
		\toprule
		Attribute $X_i$ & Total classes and their relation to original attribute values \\
		\midrule
		Female & Binary by default \\
		Misdemeanor & Binary by default \\
		African-American  & Binary by default \\
		Age below twenty five & Binary by default \\
		Number of prior crimes & 4 classes corresponding to the bins $0, [1,5], [6,10], > 10$ \\
		\bottomrule
	\end{tabular}
	\label{tab:pp_attr_3}
\end{table}

\end{document}